\documentclass[letterpaper, 10 pt, journal, twoside]{IEEEtran}

\usepackage{graphicx} 
\usepackage{hyperref}
\usepackage{comment}
\begin{document}

\title{Pinto: A latched spring actuated robot for jumping and perching}
\author{{Christopher Y. Xu$^1$, Jack Yan$^2$, and Justin K. Yim$^2$}

\thanks{$^{1,2}$The authors are with the $^1$Department of Electrical and Computer Engineering and the $^2$Department of Mechanical Science and Engineering at the University of Illinois at Urbana-Champaign, USA. 

\texttt{cyx3@illinois.edu},
\texttt{jkyim@illinois.edu}}
}
\date{September 2024}
\maketitle

\begin{abstract}
Arboreal environments challenge current robots but are deftly traversed by many familiar animal locomotors such as squirrels. We present a small, 450 g robot ``Pinto'' developed for tree-jumping, a behavior seen in squirrels but rarely in legged robots: jumping from the ground onto a vertical tree trunk. We develop a powerful and lightweight latched series-elastic actuator using a twisted string and carbon fiber springs. We consider the effects of scaling down conventional quadrupeds and experimentally show how storing energy in a parallel-elastic fashion using a latch increases jump energy compared to series-elastic or springless strategies. By switching between series and parallel-elastic modes with our latched 5-bar leg mechanism, Pinto executes energetic jumps as well as maintains continuous control during shorter bounding motions. We also develop sprung 2-DoF arms equipped with spined grippers to grasp tree bark for high-speed perching following a jump.
\end{abstract}


\section{Introduction}

We lack crucial data on forest health and biodiversity, because they are difficult to access and monitor. Currently, it is labor-intensive to diagnose the distribution of pests and diseases in order to protect agriculture and forestry \cite{Plantpests}. Still, collecting data to improve our understanding of forests has important applications in informing conservation policy, biology research, and managing a sustainable economy dependent on natural resources \cite{Xprize}. 

Robots have been developed to navigate arboreal environments in place of humans, taking the form of small drones \cite{aucone2023}, legged climbers \cite{haynes2009, Treebot}, and novel designs combining aerial and tethered movements \cite{Avocado}. These monitoring tasks require robustness in an unpredictable setting, long range, and ability to carry a small payload like sensors, computation, and wireless communication. For larger payloads, climbing robots work well to traverse steep and uneven surfaces, such as rock cliffs \cite{Loris} and telephone poles \cite{haynes2009}.

For traversing gaps and longer range, there is a large body of work applying drones to monitoring tasks; however they are fragile, have limited flight time, and need to avoid collisions in dense natural environments. Farinha et al. developed an UAV designed to launch a dart containing wireless sensors into tree bark, accessing locations unavailable to the main vehicle \cite{farinha2020}. Others developed bio-inspired grippers for drones to land reliably on large branches \cite{meng2022, roderick2021}. Aerial vehicles can be undesirable in certain applications due to noise and the downwash airflow necessary for lift that can disturb the environment being measured. Fast-spinning propellers may pose a safety risk too, and lightweight vehicles often cannot apply high manipulation forces. 

\begin{figure}[t]
    \centering
    \includegraphics[width = \columnwidth]{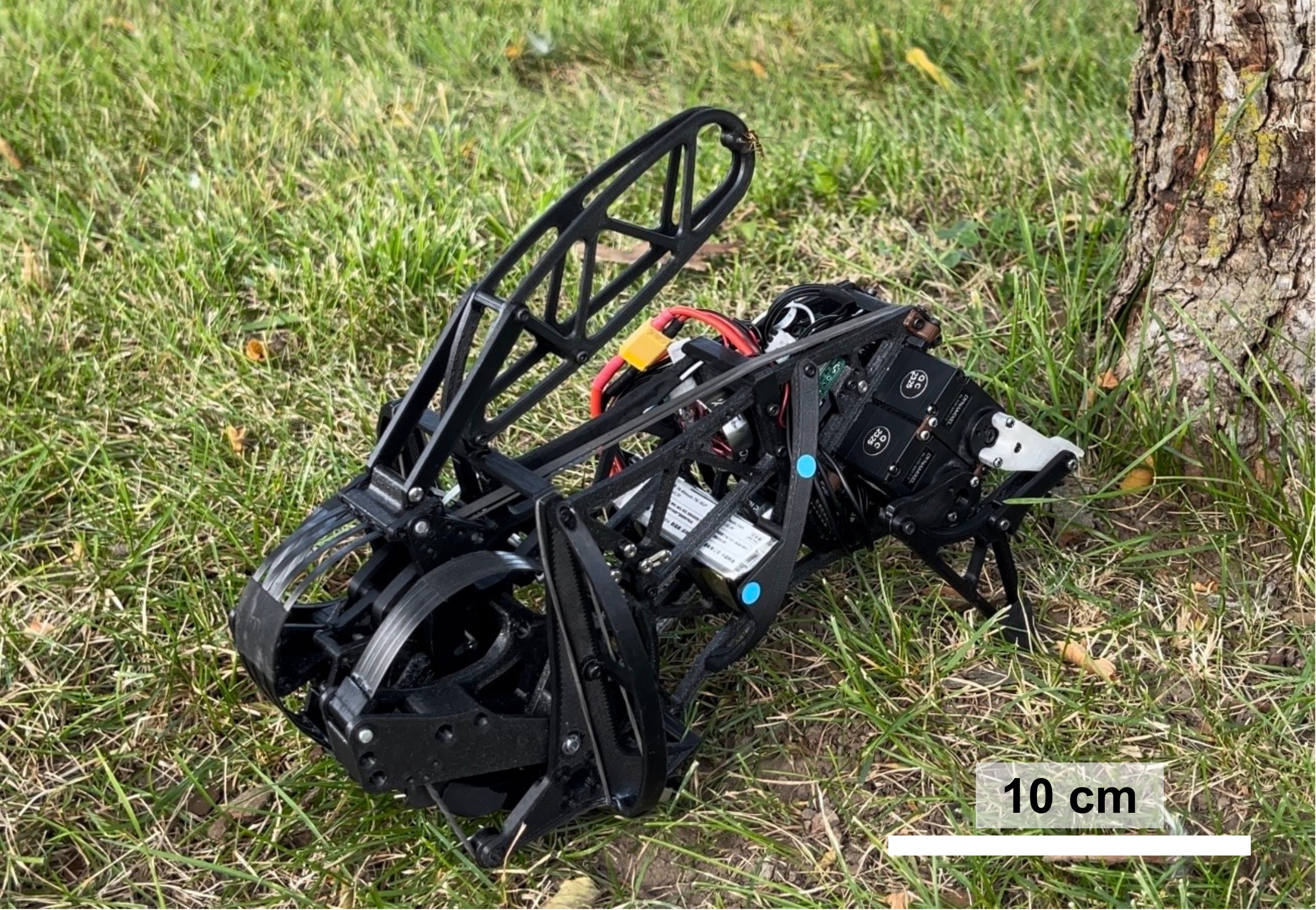}
    \caption{Pinto, a squirrel-sized robot that can jump onto a tree trunk. \\ Video: \url{https://youtu.be/-q4JTjmDX6k}}
    \label{fig:Pinto_pic}
    \vspace{-1.25em}
\end{figure}

\begin{figure}[b]
    \centering
    \includegraphics[width=\columnwidth]{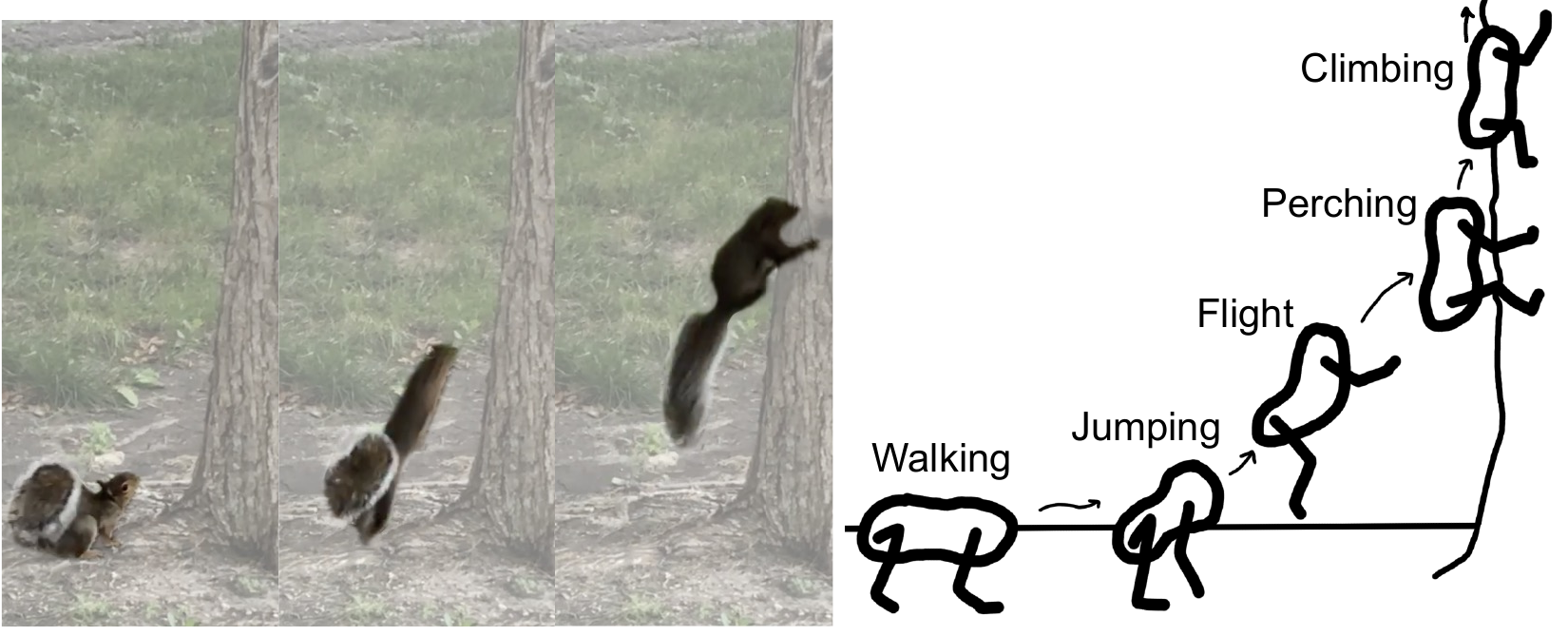}
    \caption{A squirrel demonstrating stages of a tree-jump maneuver.}
    \label{fig:squirrelpotato}
    \vspace{-1.5em}
\end{figure}

We can trade the efficiencies, speed, range, and obstacle clearance of aerial vehicles for the higher payload of agile jumping robots. Legged robots can move through tighter spaces, have longer runtimes, and produce less disturbance on their surroundings.

In the animal world, squirrels are able to navigate arboreal environments extremely well, and jumping is great for traversing between distant angled surfaces like branches in a way that static climbers do not achieve \cite{hunt2021}, in cluttered spaces difficult for drones. In addition to jumping, squirrels are also able to bound on flat ground and climb vertically. 

We aim to achieve a set of behaviors we call tree-jumping (Fig. {\ref{fig:squirrelpotato}}), whose possibility squirrels have demonstrated:
walking/bounding, jumping, flight, perching, and climbing.

In this work our prototype ``Pinto'' achieves the first four elements of tree-jumping, and to our knowledge is the first to do so without aerodynamic forces like those of flying.

We anticipate that a squirrel-sized legged mobile robot would be well-matched for navigating arboreal environments. Pinto weighs 450 grams, within the 400 to 600 gram weight range of the eastern gray squirrel (\textit{Sciurus carolinensis}), a common tree squirrel in the Midwestern United States \cite{uhlig1955}. 

Smaller size has the advantage of increasing survival during impacts from free fall and inadvertent contact when running \cite{jayaram2018}. Above an upper limit of about 1 kg, animals suffer from plastic deformation, suggesting injury. Animals below this limit may use mechanically mediated control, where impacts help create the intended motion at high speed and their active control can be less precise. For a mobile robot in a cluttered natural environment, the ability to survive impacts will help it travel faster and operate more reliably.

Additionally, small size reduces disturbance on the environment it intends to monitor compared to existing quadrupeds such as the Boston Dynamics Spot \cite{Spot}, but is still large enough to carry its own computer, wireless communication, and battery suitable for autonomous operation at long range, unlike existing insect-scale robots \cite{helbling2018}.

\section{Background}
Many existing quadrupeds achieve impressive agility using direct drive or quasi-direct drive actuators, often brushless DC motors, coupled with a low gear reduction to obtain high power with a low reflected inertia, allowing for impact absorption and estimation of contact forces using only motor current sensing \cite{seok2013, katz2019, Doggo, kenneally2016, spacebok}. As we scale down the legged robot, obstacles remain the same height but the robot’s relatively shorter legs must still propel the robot to clear them. It becomes difficult to generate the same jumping height by simply scaling down existing quadruped designs. 

Jump energy can be approximated by the work done by the motor over the stroke of the leg. We can then calculate the jump height, assuming a constant force $F$ over the stroke of the leg $L$, propelling a mass $m$. \cite{wensing2017} found that motor torque theoretically scales with the square of the gap radius r$_g$, closely related to the motor diameter. If dimensions of the robot are scaled uniformly, the torque would vary with length squared, leg stroke linearly with length scale, and mass with the length cubed. In this simple model, the length terms cancel out and predict a consistent jump height across length scales. In practice however, motors perform worse at smaller scales. In off-the-shelf frameless brushless motors, where there is minimal external structure not contributing to torque, \cite{wensing2017} found that torque density scales with $r_g^{0.8}$, where $r_g$ is the gap radius. In a review of 322 RC servos and 175 motors from 0.3 W to 250 W, \cite{dermitzakis2011} found that most commercially available light-duty motors have torques that scale proportionally with mass. These findings suggest that if motor mass scales with the cube of some characteristic length, then real motors have torques that roughly scale between $L^{2.2}$ and $L^3$. In an older study of small-scale electric motors, it was reported that torque scales with $L^{3.5}$ \cite{Alice}. This more aggressive scaling of real motors would positively correlate jump height and length scales, decreasing the expected jump height of small direct or quasi-direct drive robots.

Additionally, scaling the robot down decreases the power we are willing to supply to the motor, even if possible. The high-torque low-speed operating region typical for legged locomotion is often constrained by heating in the stator coils. In \cite{wensing2017}, a study of a catalog of frameless motors found that torque production efficiency (torque squared per unit ohmic power loss) scales with $r_g^{4.1}$, showing relatively higher heat production in smaller motors. With smaller robots using smaller radius motors and equipped with less battery energy storage due to packaging constraints, the actuators must budget more mass for higher gear reductions in order to achieve the necessary acceleration without overheating or excessively draining the battery. Higher reductions decrease the jump performance by increasing mass, decreasing maximum speed, and increasing frictional losses.

In nature, biological muscle also performs worse at jumping at smaller scales due to decreased force production at higher strain rates \cite{sutton2019}. Smaller animals can nevertheless achieve high jumps with the use of latch-mediated spring actuation (LaMSA) to amplify the instantaneous jump power beyond that of the muscle \cite{patek2023}. 

Many existing small jumping robots use a LaMSA mechanism to load up a spring in parallel with the leg \cite{brown1998, stoeter2006, Grillo, kovac2008, zaitsev2015, hawkes2022}. A winch or cam with high mechanical advantage allows low-power motors to contribute to spring energy over a long period of time, then released with a latch for impressive jumps. In \cite{hawkes2022}, a jump of over 30 meters is accomplished by a combination of work multiplication and careful design of the elastic elements. However, parallel elastic jumping robots previously mentioned lose much of their control of the end effector during the jump since the latch disconnects the motor to release the spring. This makes them inefficient to quickly retract and extend for other desired behaviors. In particular, we would like to retract and position the leg after the jump to perch onto the side of a tree trunk.

The Salto jumping monoped uses series-elastic power modulation instead. In the energy storage phase, the leg linkage produces a low mechanical advantage between the series spring and the foot, allowing the motor to wind up the spring without moving the foot due to the weight of the robot. When the forces are sufficiently high, the leg linkage starts to move and transitions to a much higher mechanical advantage, releasing the spring potential energy with power greater than that of the motor alone \cite{haldane2016}. After the foot leaves the ground, the robot can quickly retract its leg without opposition from the series-elastic (rather than parallel-elastic) spring, allowing for agile maneuvers involving position control of the foot, such as wall jumping. \cite{hong2020} uses another latch-free mechanism introducing a spring additional to the robot weight to increase the energy storage.

Our robot aims to perform climbing and walking behaviors as well as jumping. These movements often require at least two degrees of freedom (DoF) in the legs, such as independently adjusting the pitch angle and extension length of the leg. \cite{brown1998} uses a separate direct drive actuator to adjust the pitch of the 1 DoF bow leg LaMSA mechanism relative to its main body. Some direct drive or quasi direct drive quadruped jumpers arrange two actuators coaxially such that both contribute to the jumping stroke if rotated in opposite directions, and adjusts the pitch if rotated in the same direction \cite{Doggo, kenneally2016, spacebok}. This example of cooperative actuation shares the load between all the actuators, improving the jumping force capability without significantly increasing the reflected inertia with the use of additional gear reduction \cite{sim2022}.

To our knowledge, there has been no previous implementation of a cooperatively actuated 2 DoF leg with latch mediated spring actuation. \cite{spacebok} includes an optional parallel elastic spring to improve jumping performance and efficiency, but does not use a latch to control energy storage and release. In the following sections, we present the prototype robot “Pinto”, developed around a twisted string actuated 2 DoF leg mechanism capable of both latched elastic jumping and stiff position control. 

\section{Methods}
\subsection{System overview}
Pinto consists of a powerful rear leg driven by two twisted string actuators, two front arms driven by five servos, and a tail used for self-righting. The system is wirelessly controlled. 

\begin{figure}[h]
    \centering
    \includegraphics[width=\columnwidth]{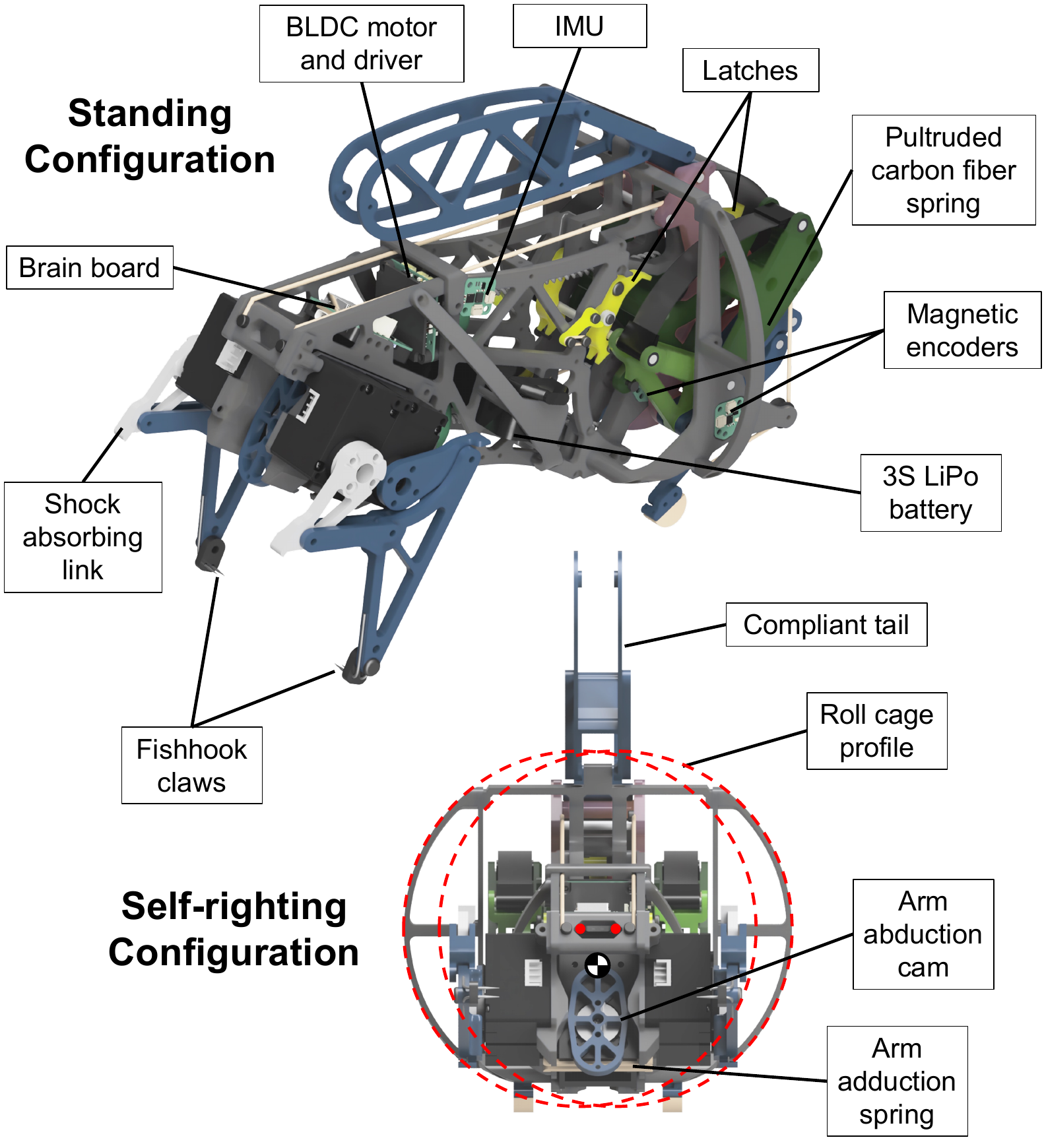}
    \caption{Diagram of Pinto. In the self-righting configuration (bottom), Pinto extends its arms forward to fit within the circular profile of its roll cage.}
    \label{fig:labeledpinto}
    \vspace{-1.25em}
\end{figure}

\begin{table}[b]
    \centering
    \caption{Pinto materials and properties}
    \begin{tabular}{|c|c|} \hline 
         Mass& 450 g\\ \hline 
         Leg stroke& 118 mm \\ \hline 
         Battery& Tattu 3S 450mAh LiPo \\ \hline 
         Leg motors& 2x 2105.5 2650KV BLDC Motors \\ \hline
         Twisted string diameter& 1.1 mm \\ \hline 
         Side springs& 3mm x 85mm x 0.5mm pultruded carbon fiber\\ \hline
         Center springs& 5mm x 85mm x 0.5mm pultruded carbon fiber\\ \hline
         Total spring energy& 2.53 J \\ \hline 
         Arm motors& 5x Dynamixel XL330-M077\\ \hline 
         Structure& 3D printed PLA, PETG \\ \hline
         Pin joints& 1/8" aluminum shafts, Igus Q bearings\\ \hline
    \end{tabular}
    \label{tab:my_label}
\end{table}

\subsection{2 DoF Latched Leg}
\begin{figure}
    \centering
    \includegraphics[width = \columnwidth]{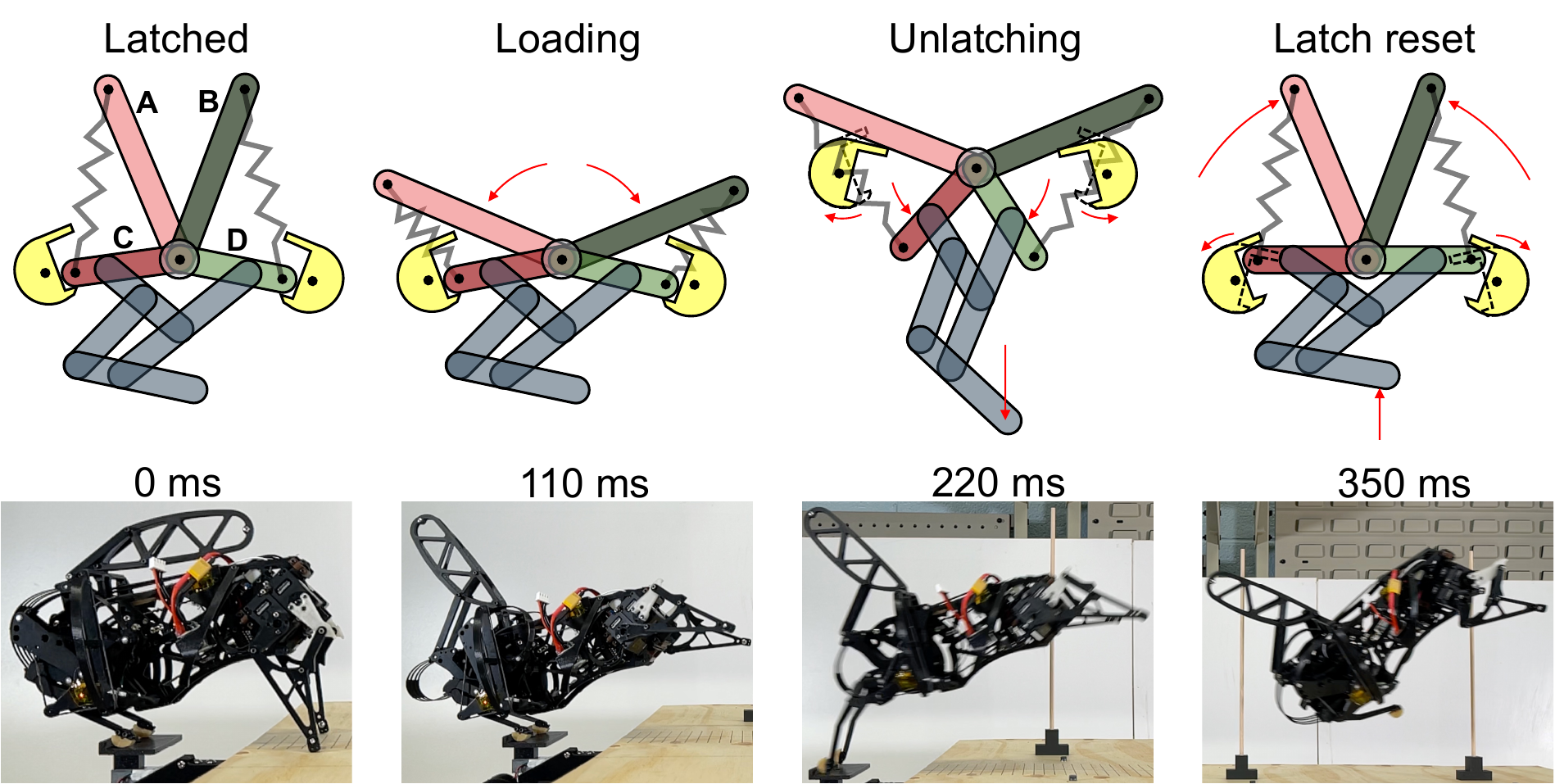}
    \caption{Time sequence of a latched jump: latches (shown in yellow) switch the system between parallel-elastic and series-elastic actuation modes. The link shapes shown here are simplified for visualization. On the robot, each link is modified for strength and denser packaging.}
    \label{fig:latch_cartoon}
    \vspace{-1.25em}
\end{figure}

We present a mechanism that can operate in either high-stiffness series-elastic or latch mediated spring actuation, allowing for both fine control and powerful jumps. A similar concept is found in the trapjaw ant, which can directly move its mandibles using its muscles, or store elastic energy in its head and rapidly release the energy for a much more powerful closing motion that can be used for locomotion \cite{sutton2022}. Fig. \ref{fig:latch_cartoon} depicts the operation of the 2 DoF LaMSA mechanism.

Links labeled A and B in Fig. \ref{fig:latch_cartoon} act as the input to the mechanism, each pushing on a spring that transfers force to links C and D. In the latched and spring loading states, links C and D are prevented from moving further, allowing the actuators to load energy into the springs. When links A and B reach a certain position they release the latches (in yellow) and the leg linkage (in blue) propels the leg downwards. After unlatching, the leg linkage is still connected to the actuator with a spring in series, allowing the leg to retract with the actuated A and B links. The latch geometry is designed such that once links C and D are sufficiently retracted, the latches reset to make it possible to compress the springs again.

The latch trigger event is solely dependent on the angle of the links, simplifying control of the spring energy release timing compared to the “dynamic catch” effect of Salto, which depends on robot weight and a race between the motor and the accelerating leg extension \cite{haldane2016, hong2020}. We place magnetic encoders to measure the angle of links C and D, and apply a proportional position controller for each degree of freedom.

By combining pultruded carbon fiber strips and latex rubber bands, the spring in \cite{hawkes2022} has a nearly constant force–displacement profile (for a “softening” nonlinear spring) that allows for motor windup up until material failure of the spring. Otherwise, an equilibrium is reached when the spring reaction force is equal to the maximum motor force \cite{sutton2019}.

\begin{figure}[h]
    \centering
    \includegraphics[width=\columnwidth]{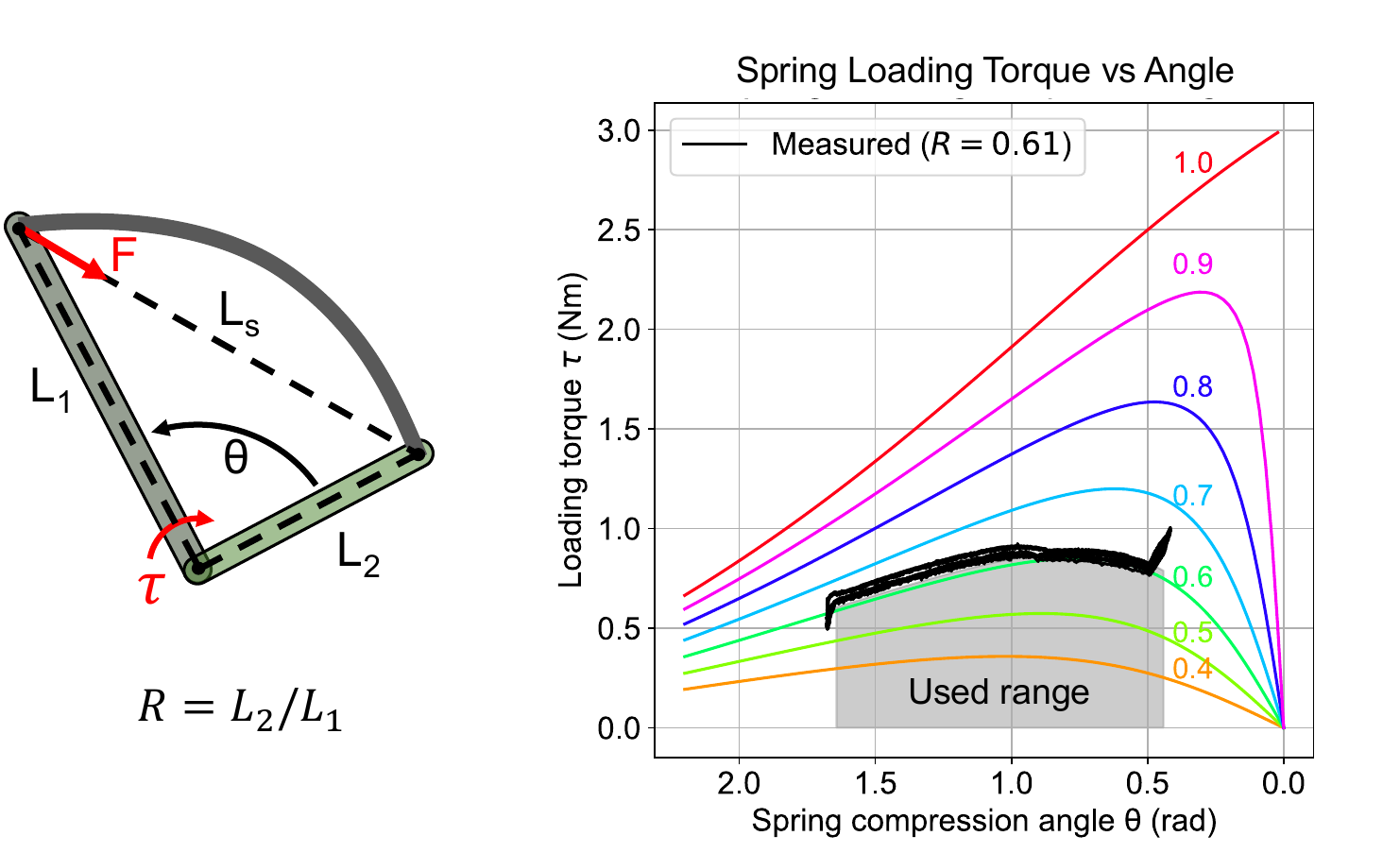}
    \caption{Experimentally measured torque vs. compression angle for one spring plotted in black and simulated torque exerted by the spring at various ratios of link length plotted in color, generated by varying $L_2$ while keeping $L_1$ and the carbon fiber strip parameters constant.}
    \label{fig:springcomplete}
\end{figure}

We use two links pinned together to load pultruded carbon fiber strips in compression and achieve a similar softening spring effect. By choosing the ratio $R$ of the output link length $L_2$ to the input link length $L_1$, we achieve a torque profile that does not increase significantly within our range of spring compression angle. Simulated springs compressed with links of different ratios R are shown in Fig. \ref{fig:springcomplete}, using an Euler buckling beam model \cite{Eulerspring}. We choose $R=0.61$ for a flatter profile that is still feasible to mechanically package.

Salto’s series-elastic power modulation requires a low stiffness spring to store enough energy for a high jump, but the low stiffness reduces the force control bandwidth of the foot, making it difficult to balance in the application of jumping between branches \cite{yim2024}. 

Pinto’s physical latch decouples the energy release from the stiffness of the spring, giving us the flexibility to choose a stiff spring to prioritize control. On the other hand, a stiffer spring requires a higher force to deflect.

To actuate the input links in the LaMSA mechanism, we use a brushless motor with a twisted string transmission for its high power and mechanical advantage. These twisted strings wrap around pulleys on links A and B. Twisted strings have been used in robotic hands for their ability to apply high forces with high speed but low torque electric motors while remaining extremely lightweight \cite{TSA, TSAhand}. Blip, a 110 kg competition battlebot, used a twisted string transmission to transfer energy from a flywheel into a mechanism to flip other robots into the air \cite{BLIP}, showing that twisted strings can be effective for highly dynamic applications. The decreasing mechanical advantage as the string contracts is well suited for accelerating a mass from rest. One limitation of the twisted string transmission is that it only applies a tension force, which must be opposed by another force for retracting the mechanism. We place relatively weaker antagonistic rubber bands to retract the leg because much higher forces are expected during leg extension than retraction. We use a section of Dyneema rope (McMaster 4377N2) due to its material durability \cite{TSAdur}, which we unbraided for smaller diameter and higher predictability \cite{TSA}.

\begin{figure}
    \centering
    \includegraphics[width = \columnwidth]{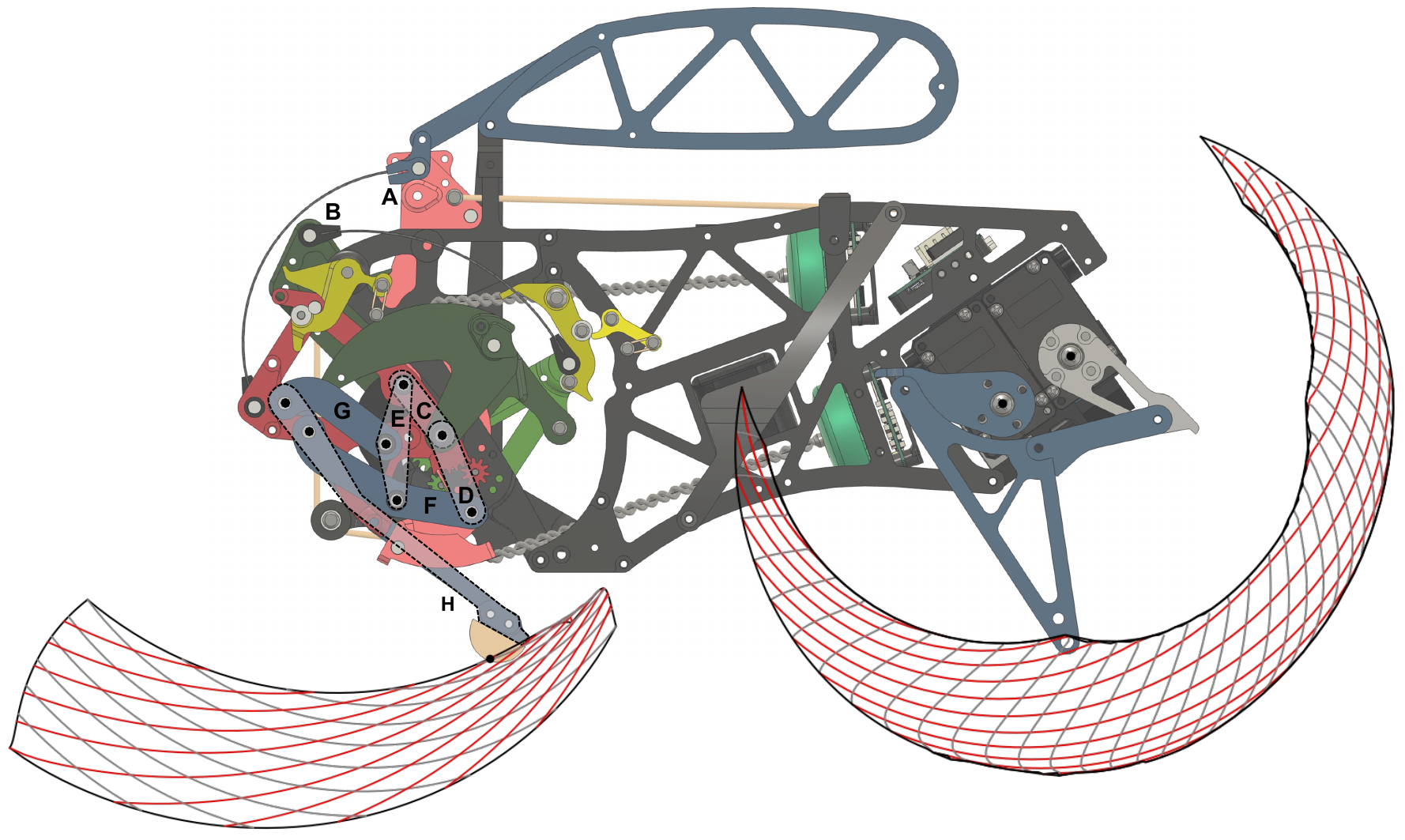}
    \caption{The workspace of the rear and front legs are shown with curves indicating the directions where both degrees of freedom are moving in the same direction (gray) and in opposite directions (red) to realize cooperative actuation \cite{sim2022}.}
    \label{fig:workspace}
    \vspace{-1.25em}
\end{figure}

The LaMSA mechanism drives links C and D, which drive the coaxial 5-bar linkage CEFD (Fig. \ref{fig:workspace}). We append two links G and H to expand the workspace while keeping the 5-bar small for easier packaging. The linkage was manually tuned to create a long leg extension trajectory when both degrees of freedom rotate simultaneously in cooperative actuation \cite{sim2022}.

\subsection{Front legs and grippers}
Pinto has 2 DoF front arms with shared abduction/adduction actuation. Each arm is tipped with 2 fish hook claws, leveraging sharp spines to grip onto asperities of hard tree bark \cite{Microspine}. One cam actuates abduction of both arms simultaneously, while adduction is provided by a rubber band, allowing for perching on tree bark without active control: during impact, the momentum of the robot pushes the arms outwards while the spring presses the claws against the bark. Similar to the rear leg, we choose a 5-bar that allows cooperative actuation \cite{sim2022} in the vertical direction for jumping, as well as in the horizontal direction for future climbing maneuvers.

\subsection{Self-righting}
 The robot automatically recovers from falls and overturning. We use a roll cage with a circular frontal profile that protects critical components from impact as well as provides passive stability on flat terrain due to the center of mass being lower than the center of the circular profile, shown in Fig. \ref{fig:labeledpinto}.

Additionally, two links coupled to the spring input link A form a tail actuated directly by the twisted string. In rough terrain where the circular roll cage is not enough to flip the robot, it may extend the tail outwards to apply a large force on the ground. While this tail produces a squirrel-like silhouette, its function is not similar to that of real squirrels' tails.

\subsection{Electrical system}
Custom BLDC motor drivers, magnetic encoders, and IMU printed circuit boards were designed for a smaller footprint and are available on Github (\url{https://github.com/qwertpas/squirrelbrain}). The boards communicate over a RS485 bus for differential noise rejection, and an ESP32 module is used to wirelessly communicate with a control station laptop.

\section{Experiments}
\subsection{Energy and time}
We evaluated the effectiveness of the latched jump behavior by comparing its jump energetics to two alternative propulsion designs: springless actuation and series-elastic actuation (SEA) without latches. With a rigid link between the twisted string output and the leg linkage replacing the springs, the jumps averaged 1.28J over 201 ms. By adding the stiff carbon fiber springs in the SEA configuration, the energy did not increase significantly and each jump took slightly longer. When adding a latch to the spring in the PEA configuration, the jump energy increased by 11\% to 1.42J, spending 246 ms (a 22\% increase in time from the springless configuration).

The SEA did not contribute to the measured jump energy compared to the springless configuration likely due to stiff springs and increased friction in the bushings. With the spring pre-stressed against the links, larger internal forces increased the load on the bushings.

\begin{figure}
    \centering
    \includegraphics[width=\columnwidth]{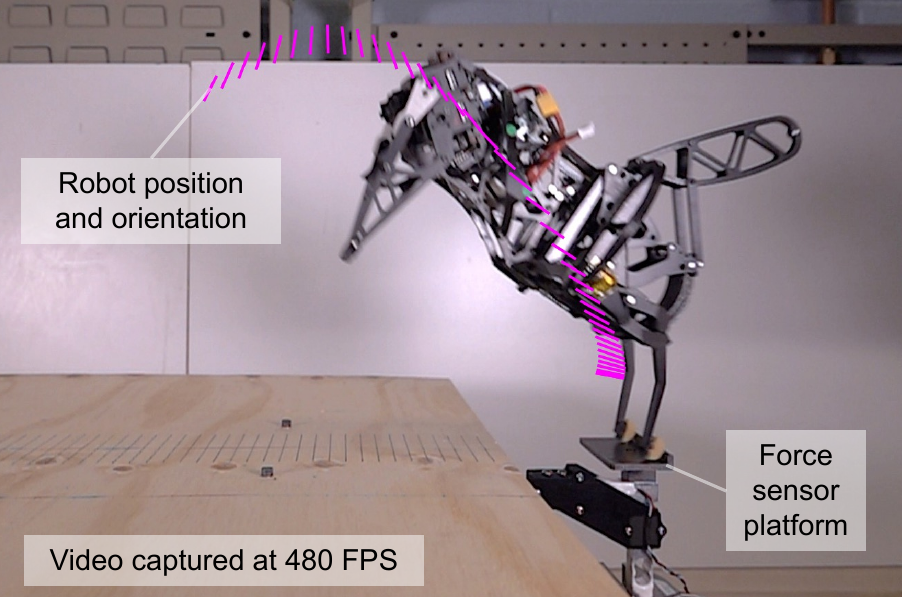}
    \caption{High speed video, onboard sensing, and a force sensor platform measure jump performance. The tracked center of mass position and robot orientation are represented by the magenta lines. We tuned the jump trajectory parameters to vertically orient the robot at the apex of the jump.}
    \label{fig:liftofftest}
    \vspace{-1.25em}
\end{figure}

\begin{figure}[b]
    \centering
    \includegraphics[width = \columnwidth]{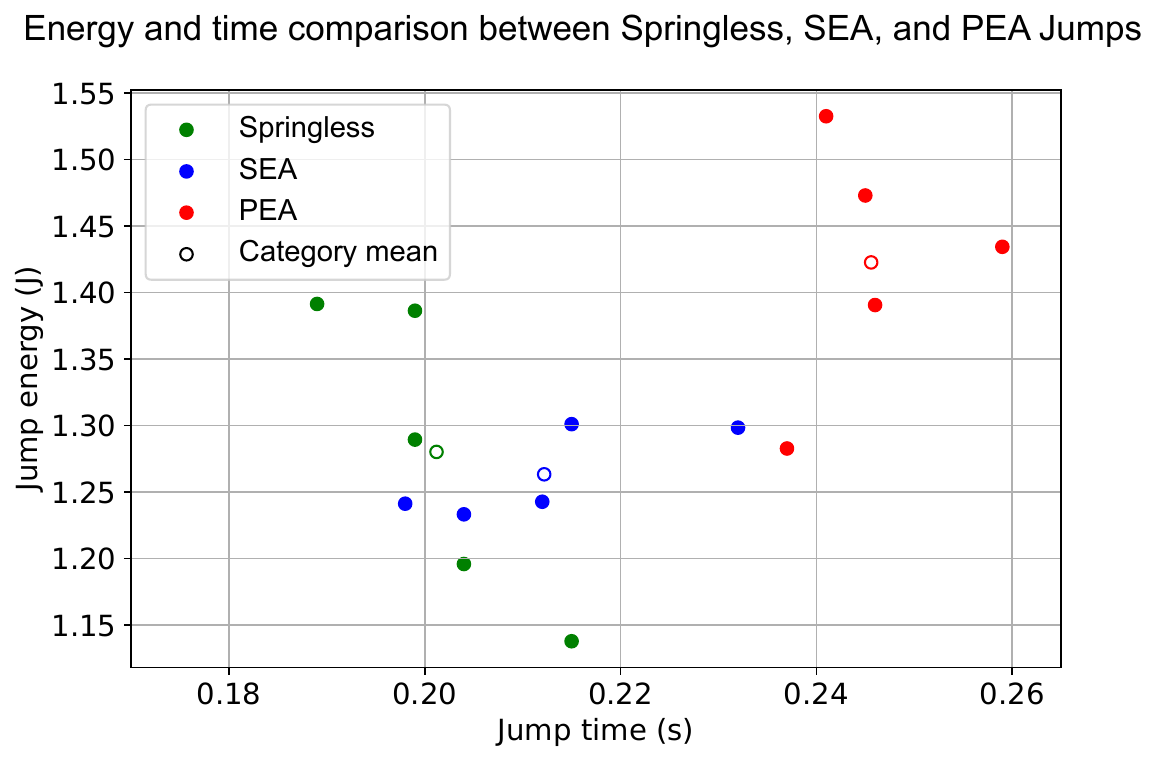}
    \caption{Jump propulsion time before takeoff and the mechanical energy immediately after takeoff were measured for 5 trials of Springless, SEA, and PEA (latched) jumps. On average, PEA took longer but gained higher energy.}
    \label{fig:energy_time_comparison_font41}
\end{figure}

\begin{figure}[t]
    \centering
    \includegraphics[width = \columnwidth]{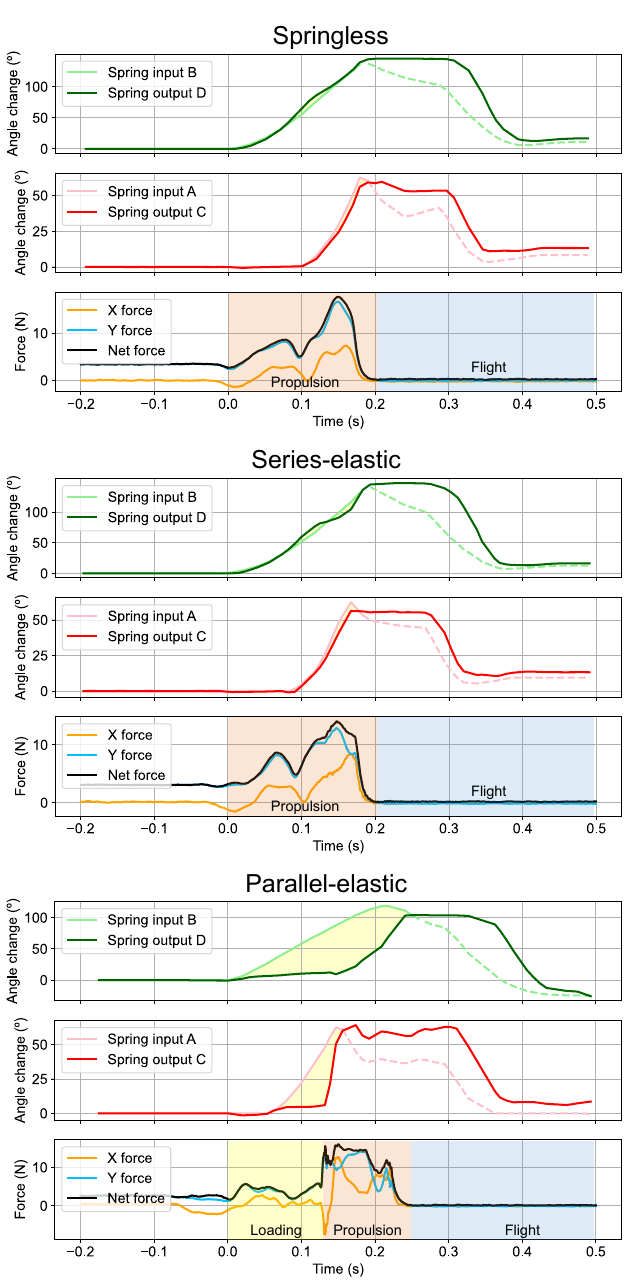}
    \caption{Link angles and ground reaction forces are plotted for springless, series-elastic, and parallel-elastic actuated jumps. Series-elastic jumps are almost indistinguishable from springless, while parallel-elastic jumping introduces a spring loading phase where the spring input and output links move relative to each other. The spring input is coupled to the motor position through the twisted string, and spring output is coupled to the foot position through the leg linkage. The area between these two curves represent spring compression and is colored in yellow. The twisted string goes slack after the spring releases energy, indicated by the dotted line.}
    \label{fig:timeplots}
    \vspace{-1.25em}
\end{figure}

We placed the rear legs on a force sensor that consists of two perpendicular load cells that measure the force in the horizontal and vertical direction. We find that the series-elastic and springless jumps have very similar force profiles consisting of two peaks. In contrast, the parallel-elastic jump has a spring loading phase without high ground forces that lasts around 130 ms, then a propulsion phase of 110 ms. The propulsion phase is shorter but high rear leg forces are maintained immediately after spring unlatch.

The average DC electrical power of the two brushless motors combined during propulsion phase was 130W for Rigid, 130W for SEA, and 180W for PEA.

\subsection{Tree jumping}

The robot performs similarly on flat floor as on uneven grassy terrain, and achieves the Launch, Flight, and Perch phases of the tree-jump behavior onto a vertical tree trunk using latched parallel-elastic actuation. Contact occurs with the tree at a center of mass height of 27 cm, while the perching results in a resting height of 22 cm (about one body length above the ground) due to front arm claws sliding to find asperities in the tree bark. 

During the flight phase, the arms and legs followed a manually tuned trajectory to the perching configuration. Inability to correct for variations in the jump resulted in a low perching success rate, however this demonstrates that a tree jump is mechanically feasible and may be actively controlled for more reliability in the future. 

\begin{figure}[h]
    \centering
    \includegraphics[width=\linewidth]{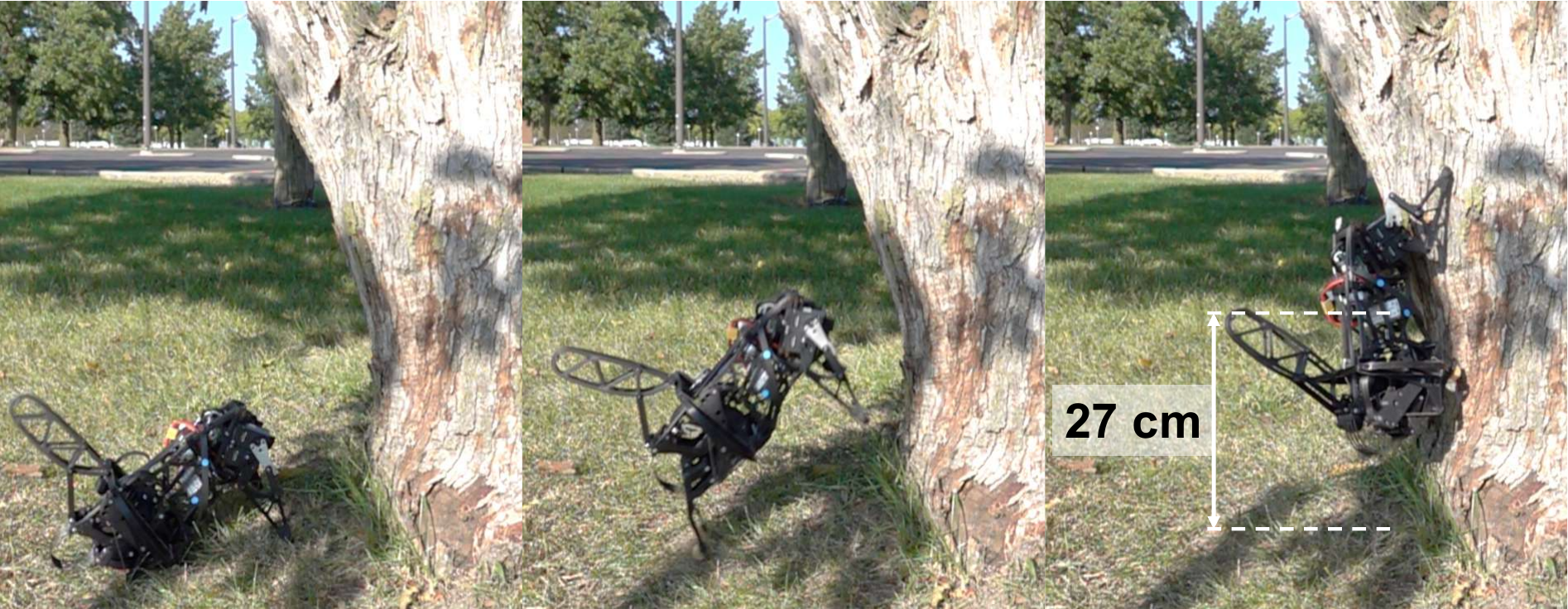}
    \caption{Pinto performing a tree jump, raising its center of mass about a body length above the ground at the time of contact.}
    \label{fig:treejump_exp}
    \vspace{-1.25em}
\end{figure}

\subsection{Other behaviors}
Operating in latched parallel-elastic or unlatched series-elastic modes allows Pinto's foot to produce both high-power jumping and detailed position control depending on the mode, improving the robot's versatility.

\begin{figure}
    \centering
    \includegraphics[width=1\linewidth]{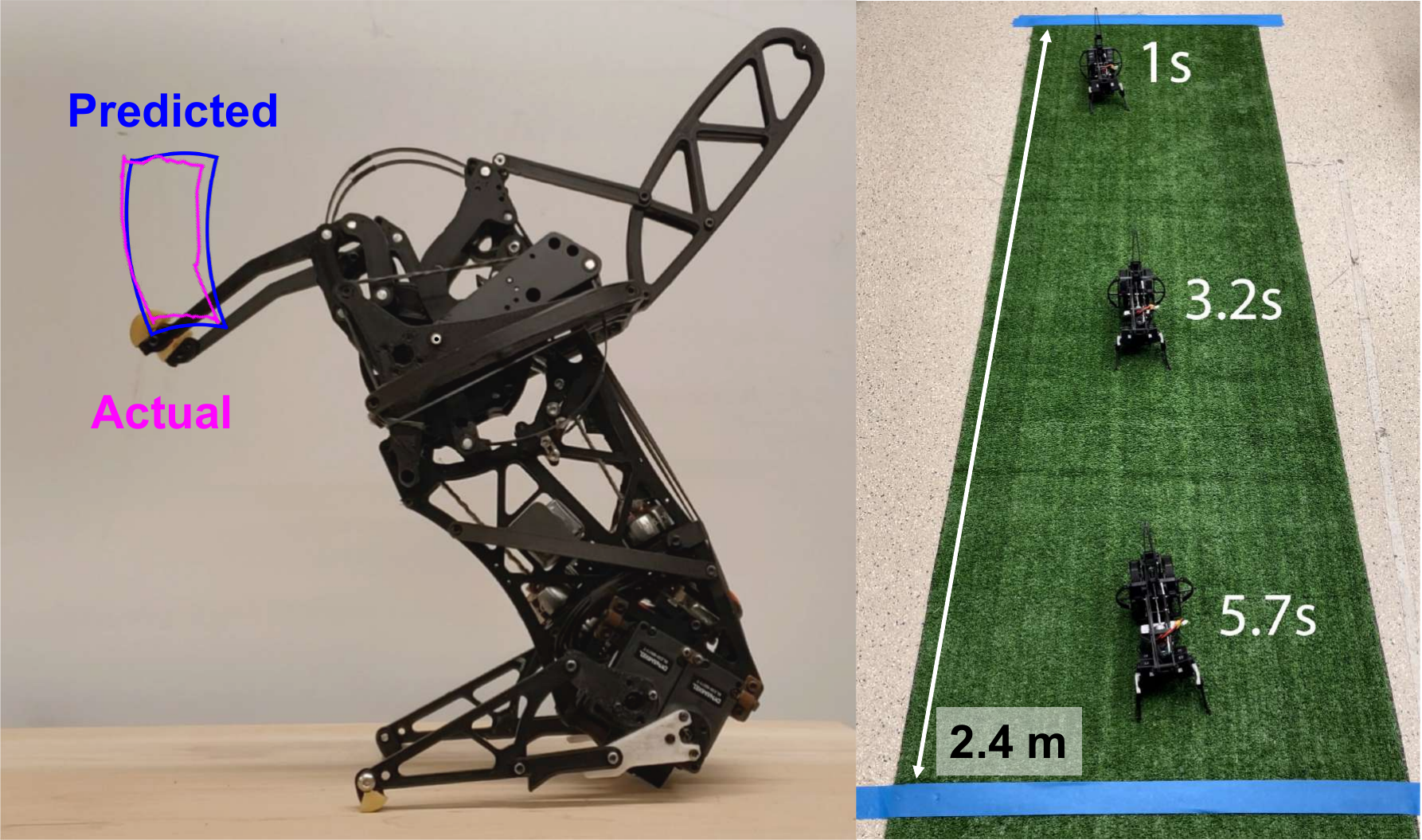}
    \caption{Other behaviors: \textit{Left}: The rear leg follows a desired trajectory. \textit{Right}: Forward bounding on an artificial turf.}
    \label{fig:otherbehaviors}
    \vspace{-1.25em}
\end{figure}

We demonstrate position control of the foot by following a closed trajectory in series-elastic mode as shown in Fig. \ref{fig:otherbehaviors}. The robot is also capable of bounding forward by chaining together multiple series-elastic jumps (reaching an average velocity of 0.39 m/s), turning using the front arms, and self-righting on both smooth and uneven terrain.

\section{Discussion}
This paper presents Pinto, a compact multi-behavior robot developed towards monitoring arboreal environments. A 2 DoF LaMSA leg mechanism with a stiff series elastic mode and a parallel elastic mode produces both precise control of the foot as well as higher jumps. Combining hobby brushless DC motors with twisted strings produces a lightweight actuator for dynamic jumping. Perching on a tree using passive compliance can be accomplished with rigid claws mounted on arms that are actuated in abduction and sprung for adduction. 

Further development may produce effective locomotion in both forest floor and arboreal environments.
Pinto demonstrates walking, jumping, perching, and even self-righting but does not demonstrate climbing on trees which could be integrated by future development.

Pinto's experimental results prove the feasibility of a multi-DoF latched series-elastic LaMSA mechanism for powerful yet controllable legged jumping at small scale.
Further performance improvement could be achieved with careful consideration of friction, impact losses, and similar implementation parameters.
For example, improvements to power consumption could be made by switching between two string twisting methods \cite{DualTSA}, or by electronically switching the motor phase winding termination \cite{Hybridwindings}, dramatically changing the motor torque constant.

Future work could also explore other regions of Pinto's multi-DoF latch paradigm.
This could include alternative release synchronization, more latched degrees of freedom, or other collaborative or antagonistic latch arrangements. For example, we originally intended to time the release of the two latches relative to each other to vary the parallel-elastic jump direction. Such mechanisms could also find applications besides jumping including manipulation or striking.

On the other hand, twisted string jump actuation proves capable on its own without latches, only 11\% less energetic than the parallel-elastic actuation in Pinto's implementation.
Springless twisted string jumpers might have lower jump energy but higher-bandwidth control and reduced mechanical complexity without elastic energy storage and latches.

Pinto's operation in both latched parallel-elastic or stiff series-elastic modes introduces a choice between operating modes for different tasks.
Future work can investigate what tasks such as climbing, manipulation, steering, or continuous hopping most benefit in each mode.

\section{Acknowledgements}
This project was funded through the 1517 Medici Project and the University of Illinois Undergraduate Research Support Grant. We thank Aleks Daeyoung Kim for his help with experiments and the beloved UIUC campus squirrel, Pinto Bean, for inspiration.

\pagebreak

\bibliographystyle{IEEEtran}
\bibliography{references.bib}

\begin{thebibliography}{10}
\providecommand{\url}[1]{#1}
\csname url@samestyle\endcsname
\providecommand{\newblock}{\relax}
\providecommand{\bibinfo}[2]{#2}
\providecommand{\BIBentrySTDinterwordspacing}{\spaceskip=0pt\relax}
\providecommand{\BIBentryALTinterwordstretchfactor}{4}
\providecommand{\BIBentryALTinterwordspacing}{\spaceskip=\fontdimen2\font plus
\BIBentryALTinterwordstretchfactor\fontdimen3\font minus \fontdimen4\font\relax}
\providecommand{\BIBforeignlanguage}[2]{{%
\expandafter\ifx\csname l@#1\endcsname\relax
\typeout{** WARNING: IEEEtran.bst: No hyphenation pattern has been}%
\typeout{** loaded for the language `#1'. Using the pattern for}%
\typeout{** the default language instead.}%
\else
\language=\csname l@#1\endcsname
\fi
#2}}
\providecommand{\BIBdecl}{\relax}
\BIBdecl

\bibitem{Plantpests}
J.~Zhang, Y.~Huang, R.~Pu, P.~Gonzalez-Moreno, L.~Yuan, K.~Wu, and W.~Huang, ``Monitoring plant diseases and pests through remote sensing technology: A review,'' \emph{Computers and Electronics in Agriculture}, vol. 165, p. 104943, 2019.

\bibitem{Xprize}
\BIBentryALTinterwordspacing
X.~Foundation, ``Xprize rainforest,'' 2024, accessed on Sept 4, 2024. [Online]. Available: \url{https://www.xprize.org/prizes/rainforest}
\BIBentrySTDinterwordspacing

\bibitem{aucone2023}
E.~Aucone, S.~Kirchgeorg, A.~Valentini, L.~Pellissier, K.~Deiner, and S.~Mintchev, ``Drone-assisted collection of environmental dna from tree branches for biodiversity monitoring,'' \emph{Science robotics}, vol.~8, no.~74, p. eadd5762, 2023.

\bibitem{haynes2009}
G.~C. Haynes, A.~Khripin, G.~Lynch, J.~Amory, A.~Saunders, A.~A. Rizzi, and D.~E. Koditschek, ``Rapid pole climbing with a quadrupedal robot,'' in \emph{2009 IEEE international conference on robotics and automation}.\hskip 1em plus 0.5em minus 0.4em\relax IEEE, 2009, pp. 2767--2772.

\bibitem{Treebot}
T.~L. Lam and Y.~Xu, ``Climbing strategy for a flexible tree climbing robot—treebot,'' \emph{IEEE transactions on robotics}, vol.~27, no.~6, pp. 1107--1117, 2011.

\bibitem{Avocado}
S.~Kirchgeorg, E.~Aucone, F.~Wenk, and S.~Mintchev, ``Design, modeling and control of avocado: A multimodal aerial-tethered robot for tree canopy exploration,'' \emph{IEEE Transactions on Robotics}, 2023.

\bibitem{Loris}
P.~Nadan, S.~Backus, and A.~M. Johnson, ``Loris: A lightweight free-climbing robot for extreme terrain exploration,'' in \emph{2024 IEEE International Conference on Robotics and Automation (ICRA)}.\hskip 1em plus 0.5em minus 0.4em\relax IEEE, 2024, pp. 18\,480--18\,486.

\bibitem{farinha2020}
A.~Farinha, R.~Zufferey, P.~Zheng, S.~F. Armanini, and M.~Kovac, ``Unmanned aerial sensor placement for cluttered environments,'' \emph{IEEE Robotics and Automation Letters}, vol.~5, no.~4, pp. 6623--6630, 2020.

\bibitem{meng2022}
J.~Meng, J.~Buzzatto, Y.~Liu, and M.~Liarokapis, ``On aerial robots with grasping and perching capabilities: A comprehensive review,'' \emph{Frontiers in Robotics and AI}, vol.~8, p. 739173, 2022.

\bibitem{roderick2021}
W.~R. Roderick, M.~R. Cutkosky, and D.~Lentink, ``Bird-inspired dynamic grasping and perching in arboreal environments,'' \emph{Science Robotics}, vol.~6, no.~61, p. eabj7562, 2021.

\bibitem{hunt2021}
N.~H. Hunt, J.~Jinn, L.~F. Jacobs, and R.~J. Full, ``Acrobatic squirrels learn to leap and land on tree branches without falling,'' \emph{Science}, vol. 373, no. 6555, pp. 697--700, 2021.

\bibitem{uhlig1955}
H.~G. Uhlig, ``Weights of adult gray squirrels,'' \emph{Journal of Mammalogy}, vol.~36, no.~2, pp. 293--296, 1955.

\bibitem{jayaram2018}
K.~Jayaram, J.-M. Mongeau, A.~Mohapatra, P.~Birkmeyer, R.~S. Fearing, and R.~J. Full, ``Transition by head-on collision: mechanically mediated manoeuvres in cockroaches and small robots,'' \emph{Journal of the Royal Society Interface}, vol.~15, no. 139, p. 20170664, 2018.

\bibitem{Spot}
\BIBentryALTinterwordspacing
B.~Dynamics, ``Spot® - the agile mobile robot,'' 2024, accessed on Sept 4, 2024. [Online]. Available: \url{https://bostondynamics.com/products/spot/}
\BIBentrySTDinterwordspacing

\bibitem{helbling2018}
E.~F. Helbling and R.~J. Wood, ``A review of propulsion, power, and control architectures for insect-scale flapping-wing vehicles,'' \emph{Applied Mechanics Reviews}, vol.~70, no.~1, p. 010801, 2018.

\bibitem{seok2013}
S.~Seok, A.~Wang, M.~Y. Chuah, D.~Otten, J.~Lang, and S.~Kim, ``Design principles for highly efficient quadrupeds and implementation on the mit cheetah robot,'' in \emph{2013 IEEE International Conference on Robotics and Automation}.\hskip 1em plus 0.5em minus 0.4em\relax IEEE, 2013, pp. 3307--3312.

\bibitem{katz2019}
B.~Katz, J.~Di~Carlo, and S.~Kim, ``Mini cheetah: A platform for pushing the limits of dynamic quadruped control,'' in \emph{2019 international conference on robotics and automation (ICRA)}.\hskip 1em plus 0.5em minus 0.4em\relax IEEE, 2019, pp. 6295--6301.

\bibitem{Doggo}
N.~Kau, A.~Schultz, N.~Ferrante, and P.~Slade, ``Stanford doggo: An open-source, quasi-direct-drive quadruped,'' in \emph{2019 International conference on robotics and automation (ICRA)}.\hskip 1em plus 0.5em minus 0.4em\relax IEEE, 2019, pp. 6309--6315.

\bibitem{kenneally2016}
G.~Kenneally, A.~De, and D.~E. Koditschek, ``Design principles for a family of direct-drive legged robots,'' \emph{IEEE Robotics and Automation Letters}, vol.~1, no.~2, pp. 900--907, 2016.

\bibitem{spacebok}
P.~Arm, R.~Zenkl, P.~Barton, L.~Beglinger, A.~Dietsche, L.~Ferrazzini, E.~Hampp, J.~Hinder, C.~Huber, D.~Schaufelberger \emph{et~al.}, ``Spacebok: A dynamic legged robot for space exploration,'' in \emph{2019 international conference on robotics and automation (ICRA)}.\hskip 1em plus 0.5em minus 0.4em\relax IEEE, 2019, pp. 6288--6294.

\bibitem{wensing2017}
P.~M. Wensing, A.~Wang, S.~Seok, D.~Otten, J.~Lang, and S.~Kim, ``Proprioceptive actuator design in the mit cheetah: Impact mitigation and high-bandwidth physical interaction for dynamic legged robots,'' \emph{Ieee transactions on robotics}, vol.~33, no.~3, pp. 509--522, 2017.

\bibitem{dermitzakis2011}
K.~Dermitzakis, J.~P. Carbajal, and J.~H. Marden, ``Scaling laws in robotics,'' \emph{Procedia Computer Science}, vol.~7, pp. 250--252, 2011.

\bibitem{Alice}
G.~Caprari, T.~Estier, and R.~Siegwart, ``Fascination of down scaling-alice the sugar cube robot,'' in \emph{IEEE International Conference on Robotics and Automation (ICRA 2000): Workshop on Mobile Micro-Robots}, 2000.

\bibitem{sutton2019}
G.~P. Sutton, E.~Mendoza, E.~Azizi, S.~J. Longo, J.~P. Olberding, M.~Ilton, and S.~N. Patek, ``Why do large animals never actuate their jumps with latch-mediated springs? because they can jump higher without them,'' \emph{Integrative and comparative biology}, vol.~59, no.~6, pp. 1609--1618, 2019.

\bibitem{patek2023}
S.~Patek, ``Latch-mediated spring actuation (lamsa): the power of integrated biomechanical systems,'' \emph{Journal of Experimental Biology}, vol. 226, no. Suppl\_1, p. jeb245262, 2023.

\bibitem{brown1998}
B.~Brown and G.~Zeglin, ``The bow leg hopping robot,'' in \emph{Proceedings. 1998 IEEE International Conference on Robotics and Automation (Cat. No. 98CH36146)}, vol.~1.\hskip 1em plus 0.5em minus 0.4em\relax IEEE, 1998, pp. 781--786.

\bibitem{stoeter2006}
S.~A. Stoeter and N.~Papanikolopoulos, ``Kinematic motion model for jumping scout robots,'' \emph{IEEE transactions on robotics}, vol.~22, no.~2, pp. 397--402, 2006.

\bibitem{Grillo}
U.~Scarfogliero, C.~Stefanini, and P.~Dario, ``Design and development of the long-jumping" grillo" mini robot,'' in \emph{Proceedings 2007 IEEE International conference on robotics and automation}.\hskip 1em plus 0.5em minus 0.4em\relax IEEE, 2007, pp. 467--472.

\bibitem{kovac2008}
M.~Kovac, M.~Fuchs, A.~Guignard, J.-C. Zufferey, and D.~Floreano, ``A miniature 7g jumping robot,'' in \emph{2008 IEEE international conference on robotics and automation}.\hskip 1em plus 0.5em minus 0.4em\relax IEEE, 2008, pp. 373--378.

\bibitem{zaitsev2015}
V.~Zaitsev, O.~Gvirsman, U.~B. Hanan, A.~Weiss, A.~Ayali, and G.~Kosa, ``A locust-inspired miniature jumping robot,'' \emph{Bioinspiration \& biomimetics}, vol.~10, no.~6, p. 066012, 2015.

\bibitem{hawkes2022}
E.~W. Hawkes, C.~Xiao, R.-A. Peloquin, C.~Keeley, M.~R. Begley, M.~T. Pope, and G.~Niemeyer, ``Engineered jumpers overcome biological limits via work multiplication,'' \emph{Nature}, vol. 604, no. 7907, pp. 657--661, 2022.

\bibitem{haldane2016}
D.~W. Haldane, M.~M. Plecnik, J.~K. Yim, and R.~S. Fearing, ``Robotic vertical jumping agility via series-elastic power modulation,'' \emph{Science Robotics}, vol.~1, no.~1, p. eaag2048, 2016.

\bibitem{hong2020}
C.~Hong, D.~Tang, Q.~Quan, Z.~Cao, and Z.~Deng, ``A combined series-elastic actuator \& parallel-elastic leg no-latch bio-inspired jumping robot,'' \emph{Mechanism and machine theory}, vol. 149, p. 103814, 2020.

\bibitem{sim2022}
Y.~Sim and J.~Ramos, ``Tello leg: The study of design principles and metrics for dynamic humanoid robots,'' \emph{IEEE Robotics and Automation Letters}, vol.~7, no.~4, p. 9318–9325, 2022.

\bibitem{sutton2022}
G.~P. Sutton, R.~St~Pierre, C.-Y. Kuo, A.~P. Summers, S.~Bergbreiter, S.~Cox, and S.~Patek, ``Dual spring force couples yield multifunctionality and ultrafast, precision rotation in tiny biomechanical systems,'' \emph{Journal of Experimental Biology}, vol. 225, no.~14, p. jeb244077, 2022.

\bibitem{Eulerspring}
A.~Klaptocz, A.~Briod, L.~Daler, J.-C. Zufferey, and D.~Floreano, ``Euler spring collision protection for flying robots,'' in \emph{2013 IEEE/RSJ International Conference on Intelligent Robots and Systems}, 2013, pp. 1886--1892.

\bibitem{yim2024}
J.~K. Yim, E.~K. Wang, S.~D. Lee, N.~H. Hunt, R.~J. Full, and R.~S. Fearing, ``Monopedal robot branch-to-branch leaping and landing inspired by squirrel balance control,'' \emph{(under review)}, 2024.

\bibitem{TSA}
I.~Gaponov, D.~Popov, and J.-H. Ryu, ``Twisted string actuation systems: A study of the mathematical model and a comparison of twisted strings,'' \emph{IEEE/ASME Transactions on mechatronics}, vol.~19, no.~4, pp. 1331--1342, 2013.

\bibitem{TSAhand}
R.~Konda, D.~Bombara, S.~Swanbeck, and J.~Zhang, ``Anthropomorphic twisted string-actuated soft robotic gripper with tendon-based stiffening,'' \emph{IEEE Transactions on Robotics}, vol.~39, no.~2, pp. 1178--1195, 2022.

\bibitem{BLIP}
\BIBentryALTinterwordspacing
S.~R. Robotics, ``Blip reveal [seems reasonable robotics],'' 2021, accessed on June 20, 2024. [Online]. Available: \url{https://www.youtube.com/watch?v=bzZufNCXaeE}
\BIBentrySTDinterwordspacing

\bibitem{TSAdur}
S.~Sadachika, M.~Kanekiyo, H.~Nabae, and G.~Endo, ``Repetitive twisting durability of synthetic fiber ropes,'' in \emph{2023 IEEE International Conference on Robotics and Automation (ICRA)}.\hskip 1em plus 0.5em minus 0.4em\relax IEEE, 2023, pp. 7412--7418.

\bibitem{Microspine}
S.~Kim, A.~Asbeck, M.~Cutkosky, and W.~Provancher, ``Spinybotii: climbing hard walls with compliant microspines,'' in \emph{ICAR '05. Proceedings., 12th International Conference on Advanced Robotics, 2005.}, 2005, pp. 601--606.

\bibitem{DualTSA}
S.~H. Jeong, Y.~J. Shin, and K.-S. Kim, ``Design and analysis of the active dual-mode twisting actuation mechanism,'' \emph{IEEE/ASME Transactions on Mechatronics}, vol.~22, no.~6, pp. 2790--2801, 2017.

\bibitem{Hybridwindings}
Y.-T. Chen, C.-L. Chiu, Y.-R. Jhang, Z.-H. Tang, and R.-H. Liang, ``A driver for the single-phase brushless dc fan motor with hybrid winding structure,'' \emph{IEEE Transactions on Industrial Electronics}, vol.~60, no.~10, pp. 4369--4375, 2012.

\end{thebibliography}

\end{document}